\documentclass[journal]{IEEEtran}

\usepackage{cite}
\usepackage{amsmath}
\usepackage{graphicx}
\usepackage{url}
\usepackage{hyperref}
\usepackage{booktabs}
\usepackage{listings}

\begin{document}

\title{Executable Archaeology: Reanimating the Logic Theorist from its IPL-V Source}

\author{Jeff~Shrager,~Ph.D.\\
Bennu Climate, Inc.\\
\texttt{jshrager@gmail.com}}

\maketitle

\begin{abstract}

The Logic Theorist (LT), created by Allen Newell, J.~C.~Shaw, and
Herbert Simon in 1955--1956, is widely regarded as the first
artificial intelligence program.  While the original conceptual model
was described in 1956, it underwent several iterations as the
underlying Information Processing Language (IPL) evolved.  Here I
describe the construction of a new IPL-V interpreter, written in
Common Lisp, and the faithful reanimation of the Logic Theorist from
code transcribed directly from Stefferud's 1963 RAND technical report.
Stefferud's version represents a pedagogical re-coding of the original
heuristic logic into the standardized IPL-V. The reanimated LT
successfully proves 16 of 23 attempted theorems from Chapter 2 of
\textit{Principia Mathematica}, results that are historically
consistent with the original system's behavior within its search
limits.  To the author's knowledge, this is the first successful
execution of the original Logic Theorist code in over half a century.

\end{abstract}


\section{Introduction}
\label{sec:intro}

In the summer of 1956, a small group of researchers gathered at
Dartmouth College for a workshop on what John McCarthy had dubbed
``artificial intelligence.'' Among them were Allen Newell and Herbert
Simon, who arrived with something none of the other attendees had: a
working AI program. While the other participants came with proposals
and ideas, Newell and Simon brought a system that actually
ran---a program that could discover proofs of mathematical
theorems.\footnote{It is worth noting that Art Samuel's
checkers-playing program existed a couple of years before LT, and it
too was heuristic. But Samuel's goal was to build a checkers player,
whereas Newell, Simon, and Shaw's explicit goal was to model human
cognition across many domains. Samuel's effort was brilliant in its
own right---he also built arguably the first machine learning program,
and the first system to learn from self-play.}

The Logic Theorist (LT)\cite{LT1956a} would ultimately prove 38 of the
first 52 theorems in Chapter~2 of Whitehead and Russell's
\textit{Principia Mathematica}~\cite{whitehead1910principia}, in some
cases finding proofs more elegant than those produced by the human
authors.  For Theorem~2.85, LT discovered a shorter, more direct proof
than the one published in \textit{Principia}.  Simon showed the new
proof to Bertrand Russell himself, who ``responded with
delight''~\cite[p.~167]{mccorduck2004machines}. Edward Feigenbaum,
then an undergraduate taking a graduate course from Simon,
recalls the moment the project's significance was first announced. As
he later told Pamela McCorduck: ``It was just after Christmas
vacation---January 1956---when Herb Simon came into the classroom and
said, `Over Christmas Allen Newell and I invented a thinking machine.'
And we all looked blank''~\cite[p.~138]{mccorduck2004machines}.

Beyond reproducing \textit{Principia} proofs, the Logic Theorist
introduced heuristic search strategies---means--ends analysis and
subgoal decomposition---derived from protocol analyses of human
problem solving. These ideas later became foundational in both
artificial intelligence and cognitive architectures such as the
General Problem Solver (GPS)~\cite{newell1959gps}, Soar, ACT, and
EPAM~\cite{feigenbaum1961epam}. LT thus sits at the very beginning of
the fusion of AI and cognitive psychology.

LT was implemented in the Information Processing Language (IPL), a
language created by the same team---Newell, Shaw, and
Simon---specifically for building cognitive
models~\cite{newell1964ipl}. IPL introduced list manipulation, dynamic
memory allocation, symbols as first-class objects, recursion, and
higher-order functions. It was explicitly a direct ancestor of
Lisp~\cite{mccarthy1960lisp}---McCarthy himself acknowledged IPL's
influence. Yet while Lisp has thrived for nearly seven decades, IPL
has been extinct since the mid-1960s; While numerous IPL-V
implementations existed in the 1960s on various machines, no
maintained implementations appear to be available in modern
environments.  The only modern attempt known to the author is
Richman's unpublished IPL-V interpreter, written in Microsoft BASIC,
which he used to run EPAM~IV in the early
1990s~\cite{richman1995epam}.\footnote{Richman wrote to the author
(personal communication, Sep~26, 2025, quoted with permission): ``I
just interpreted IPL-V in Microsoft Basic using the IPL-V manual that
I found at the CMU library.''}

This paper describes the construction of a complete IPL-V interpreter
in Common Lisp, and the faithful reanimation of the Logic Theorist
from code transcribed directly from Stefferud's 1963 RAND technical
report~\cite{stefferud1963lt}. More than 99\% of the LT code running
in this reconstruction is the original code from that report. The
reanimated LT proves 16 of 23 attempted theorems from
\textit{Principia Mathematica}, results that are historically
consistent with the original LT's known behavior.

This work illustrates a methodological approach that might be called
\emph{executable archaeology}: the study of historically significant
computational systems through faithful reconstruction and execution of
their original code. Early AIs were not merely theoretical proposals
but functioning programs whose detailed behavior cannot be fully
understood from published descriptions alone. Reanimating such systems
provides a new empirical window into the origins of artificial
intelligence.

In prior work, a team including the present author restored
Weizenbaum's original ELIZA to operation on a reconstructed CTSS
running on an emulated IBM 7094~\cite{lane2025eliza}. Although the
Logic Theorist and ELIZA addressed very different problems---formal
theorem proving versus natural-language conversation---both programs
emerged from the same symbolic programming lineage originating in IPL
and later realized in Lisp and related list-processing systems. Each
can be understood as a rule-based control system operating over
symbolic list structures. However, whereas the ELIZA work rested on a
stack that emulated vintage hardware and operating systems, the LT
reanimation emulates the abstract IPL-V machine.\footnote{Ironically, in
IPL's direct descendant language, Common Lisp.}

\section{Historical Background}
\label{sec:history}

\subsection{The Origins and Versions of the Logic Theorist}

The story of the Logic Theorist begins in the early 1950s at the RAND
Corporation.  Herbert Simon and Allen Newell, collaborating with
J.~C. (Cliff) Shaw, sought to simulate complex thought by manipulating
symbols.  However, the history of the program is characterized by a
series of shifting specifications across different versions of the
Information Processing Language.

The system described in the famous 1956 RAND report (P-868) was a
``paper'' version, often referred to as IPL-I or the ``Logic
Language'' (LL).  At the time of the Dartmouth workshop in the summer
of 1956, the program was not yet running on a machine; Newell and
Simon noted it was ``essentially specified'' and that hand simulations
suggested it would prove the 60-odd theorems of Chapter 2 of
\textit{Principia Mathematica}.  Before the program ran on any
machine, Newell and Simon famously simulated parts of it by hand using
``the human computer''—family members and students acting as program
components.

It was not until the implementation of IPL-II on the JOHNNIAC in late
1956 and early 1957\cite{gruenberger1968johnniac} that
machine-executable proofs were generated.  The widely cited result
that LT proved 38 of the first 52 theorems in
\textit{Principia}—including the discovery of a more elegant proof for
Theorem 2.85 than the one produced by Whitehead and Russell—stems from
these 1957 runs.

The code used in this reanimation, transcribed from Stefferud's 1963
memorandum, represents a third distinct phase.  Stefferud's version
was a pedagogical re-coding of the original heuristic logic into
IPL-V, the first version of the language made widely available for
public use.  Thus, while our reanimation is a faithful execution of
original code published by the RAND team, it is a restoration of
the 1963 ``modernized'' version of the 1956 conceptual model.

\begin{table}[h]
\centering
\caption{Evolution of the Logic Theorist and IPL}
\label{tab:lt_versions}
\begin{tabular}{@{}llll@{}}
\toprule
\textbf{Year} & \textbf{Version} & \textbf{Platform} & \textbf{Historical Milestones} \\ \midrule
1956 & IPL-I (LL) & Manual & Conceptual spec; Hand-simulated only. \\
1957 & IPL-II & JOHNNIAC & Proved 38 of 52 \textit{Principia} theorems. \\
1958 & IPL-IV & IBM 704 & Internal RAND use; transitioning to IPL-V. \\
1963 & IPL-V & Various & Stefferud's pedagogical LT implementation. \\
2025 & IPL-V & Lisp/LLM & Reanimation using Stefferud's source code. \\ \bottomrule
\end{tabular}
\end{table}

\subsection{IPL: The Invention of Symbol- and List-Processing}

To implement LT, Newell, Shaw, and Simon needed a programming language
suited to symbolic reasoning.  No such language existed, so they
created the ``Information Processing Language.''  As noted, IPL went
through several iterations.  While the earliest conceptual versions
(IPL-I and IPL-II) powered the 1956 hand simulations and subsequent
JOHNNIAC runs, IPL-V, released in the early 1960s~\cite{newell1964ipl}, was the version
that became the definitive public standard.

IPL-V runs on an abstract machine organized around \textit{cells}.
Each cell has an associated symbol. There is no single distinguished
stack; every cell is potentially a stack of symbols that can be pushed
and popped. Symbols are central to IPL-V: they are first-class
objects, and every cell, data structure, routine, and control element
is addressable by a symbol (its name). A set of control cells
(H0--H5) govern execution: H0 serves as the parameter and result
``communication'' cell, H1 serves as both program counter and call
stack, H2 is the free list for dynamic memory allocation, and H3 is
the cycle counter. Ten working cells (W0--W9) provide scratch
storage. Everything in IPL-V---data, routines, even the program
itself---is a list of cells, and thus also a list of symbols. Routines
are invoked by pushing their symbol (name) onto H1. The IPL machine
interprets the operation indicated in that cell, and then continue to
whatever is the indicated next cell. Elegantly, because H1 is both
program counter and call stack, no special machinery is required to
operate the machine aside from basic stack manipulation
instructions.

The language's built-in operations are called J-functions (because
their names begin with ``J'': J0, J1, J2, and so on). There are
approximately 150 of these, implementing operations from basic list
manipulation to ``generator'' (iterator) protocols. In the original
IPL-V, many J-functions were themselves written in IPL, although some
were implemented at real machine-level.

As mentioned above, IPL was an explicit direct ancestor of
Lisp~\cite{mccarthy1960lisp}. It introduced list and symbol
manipulation, the idea that programs are themselves lists, dynamic
memory allocation, recursion, and higher-order functions. Having now
spent months implementing an IPL-V interpreter, I can report that IPL
had essentially \textit{everything} that Lisp has, with two critical
exceptions: homoiconic syntax, and automatic garbage collection. Where
Lisp unified code and data in a single, elegant notation, IPL code was
expressed in an assembly-language style which, while powerful, was far
more difficult to read and write. And where Lisp introduced garbage
collection to automate memory reclamation, IPL required explicit
management of free storage.

It is worth noting that IPL-V programs could, in principle, modify
their own code at runtime---the same self-modification capability for
which Lisp is celebrated. Because IPL-V routines are simply lists, and
lists can be manipulated by IPL-V programs, nothing prevents a program
from rewriting its own procedures. In practice this would be far more
cumbersome than in Lisp, precisely because of the lack of homoiconic
syntax, but the capability was there. McCarthy's genius was not in
inventing list processing \textit{de novo}, but in finding the
syntactic and runtime forms that made these ideas---ideas that Newell,
Simon, and Shaw had already developed---elegant, accessible, and
practical. 

\section{Source Materials and Transcription}
\label{sec:sources}

Unlike the ELIZA reanimation, which began with the dramatic discovery
of original source code printouts in Weizenbaum's archives at
MIT~\cite{lane2025eliza}, the source materials for the Logic Theorist
were already publicly available. Two documents were indispensable:

\begin{enumerate}
\item \textbf{Stefferud (1963)}, \textit{The Logic Theory Machine: A
  Model Heuristic Program}, RAND Corporation memorandum
  RM-3731~\cite{stefferud1963lt}. This report contains a complete
  listing of the LT program in IPL-V, along with detailed descriptions
  of every aspect of LT's workings, including numerous flow charts.

\item \textbf{Newell et al.\ (1964)}, \textit{Information Processing
  Language-V Manual}, 2nd edition~\cite{newell1964ipl}. The definitive
  reference for IPL-V, including the semantics of all J-functions,
  cell structure conventions, and the generator protocol, as well as
  numerous student exercises and sample programs.
\end{enumerate}

Both documents proved to be exceptionally well-written. There were,
however, occasional ambiguities---places where the specification
assumed familiarity with conventions of 1960s computing (such as BCD
character encoding) that are no longer common knowledge.

The LT code was transcribed from Stefferud's report into a Google
spreadsheet by the author and Anthony Hay. Subsequently, Rupert Lane
identified several transcription errors using his ``gridlock''
tool~\cite{lane2024gridlock}, software designed to accurately extract
tabular code from PDF documents. Despite multiple people putting
multiple sets of eyes on the code, a surprising number of small
transcription errors persisted through numerous rounds of
review.\footnote{I probably shouldn't have been surprised by this
given the number of typos that have persisted through numerous rounds
of review of this paper!}

The transcribed code was converted into an S-expression format
(\texttt{.liplv} files) suitable for loading by the Lisp-based
interpreter. Conveniently, Lisp comment characters work in these
files, allowing corrections and annotations to be documented inline.

\section{A New IPL-V Interpreter in Lisp}
\label{sec:interpreter}

I implemented a new IPL-V interpreter to execute the Logic Theorist
listing. The interpreter implements the symbolic data structures and
execution model described in the IPL-V manual and supports the
primitives required by the program. Although the earliest versions of
the Logic Theorist ran on the RAND JOHNNIAC, by the time the Stefferud
listing, used here, was published (1963) IPL-V had already been
implemented on multiple machines. Accordingly, this reconstruction
executes the program within a machine-independent IPL-V interpreter
rather than attempting to reproduce a specific JOHNNIAC hardware
environment. The full interpreter and transcription of the Logic
Theorist code are available in the repository associated with this
paper.

I chose to emulate the IPL-V abstract machine, rather than emulating
the original hardware environment (as was done for ELIZA on the IBM
7094 and CTSS~\cite{lane2025eliza}), for several reasons. First, as
mentioned just above, the code I am using is IPL-V, and so is not
JOHNNIAC specific. Additionally, no thorough JOHNNIAC emulator exists,
and in any case we do not have the JOHNNIAC IPL code. There does exist
an IBM 7094 emulator that includes an IPL-V interpreter running under
a variant of the IBSYS batch operating system \cite{ibsys}, and this
emulator includes demo scripts for IPL-V. However, this IPL-V
interpreter is a binary with no source code and no documentation. I
attempted to run LT on this platform but it failed early in execution
with no useful information about the cause. Because there is no source
code, this failure was essentially undebuggable.

More fundamentally, however, one of my central goals in undertaking
this project was not merely to get LT to run, but to
\textit{elucidate} IPL-V---to develop a deep understanding of the
language, the machine model, and the infrastructure of the
period. Writing my own interpreter was essential to that goal. IPL-V
is defined as an abstract machine specification, not as a program for
a specific computer; it was implemented on many different machines
through the mid-1960s before being supplanted by Lisp.\footnote{In
addition there is a pleasing circularity in implementing
IPL-V---Lisp's direct ancestor---in Lisp itself.}

My interpreter (\texttt{iplv.lisp}) implements the complete IPL-V
abstract machine as described in the 1964 manual: the cell table, the
control cells H0--H5, the working cells W0--W9, and the push-down
mechanisms. Lists are stored as linked chains of cells, exactly as
specified. It is worth noting that the push-down (stack) mechanism is
itself implemented as linked lists within the IPL-V system. Push and
pop operations, which appear to be hardware primitives of the abstract
machine, are internally built on the same list mechanism that
underlies everything else in IPL-V, including data structures,
routines, and IPL programs themselves; like Lisp, IPL-V is lists all
the way down.

The core of my interpreter is a function, \texttt{ipl-eval}, that
fetches and executes instructions by executing the list named in the
program counter (H1). My IPL evaluator is re-entrant --- a capability
trivially implementable in Lisp --- supporting the recursive calls that
IPL-V requires. The J-functions are implemented as Lisp
functions. While the interpreter implements the full IPL-V machine,
only the subset of J-functions required by LT (and a few test
programs) is currently implemented, amounting to several dozen of the
approximately 150 defined in the manual.

A complete run of LT attempting proofs of 23 theorems is available at
\cite{runoutput}. This takes approximately 574,000 IPL machine cycles
and creates over 576,000 symbols\footnote{Yeah, I was surprised
too. It's really just a coincidence!}. It finishes in a few seconds on
a typical modern laptop. On the JOHNNIAC at RAND, in the mid 1950s,
this likely would have required hours.

\section{What the Logic Theorist Proves}
\label{sec:results}

The Logic Theorist attempts to prove theorems in propositional logic
using the five axioms of \textit{Principia Mathematica}, Chapter~2.
The notation used by the original authors is given in
Table~\ref{tab:notation}.

\begin{table}[h]
\centering
\caption{Logical notation used by the Logic Theorist}
\label{tab:notation}
\begin{tabular}{cl}
\toprule
\textbf{Symbol} & \textbf{Meaning} \\
\midrule
\texttt{I} & Implication ($\rightarrow$) \\
\texttt{V} & Disjunction ($\lor$) \\
\texttt{*} & Conjunction ($\land$) \\
\texttt{-} & Negation ($\lnot$) \\
\texttt{=} & Biconditional ($\leftrightarrow$) \\
\texttt{.=.} & Definitional equality \\
\bottomrule
\end{tabular}
\end{table}

LT employs several proof methods:

\begin{itemize}
\item \textbf{Substitution}---substitute variables in an axiom or
  previously proved theorem to match the goal.
\item \textbf{Detachment} (modus ponens)---from $A \rightarrow B$ and
  $A$, conclude $B$.
\item \textbf{Forward chaining}---apply substitution and detachment
  working forward from known theorems.
\item \textbf{Backward chaining}---work backward from the goal,
  seeking premises that would yield it.
\item \textbf{Sublevel replacement}---replace a subexpression using a
  known equivalence.
\end{itemize}

Each proof attempt is bounded by a settable effort limit in terms of
IPL machine cycles. (In all below examples this limit is set to 20,000
cycles.) This is a soft cap: it prevents the system from
\textit{beginning} a new subproblem exploration once exceeded, but
does not interrupt a search in progress.

Table~\ref{tab:axioms} shows the axioms and definitions that are
initially provided to LT for the runs described in the present
paper. Importantly, as LT proves a given theorem, that theorem is
treated as an axiom and can be used to support subsequent
proofs. Thus, LT can be said, in a sense, to learn!

\begin{table}[h]
\centering
\caption{Axioms and definitions}
\label{tab:axioms}
\begin{tabular}{ll}
\toprule
\textbf{Name} & \textbf{Axiom/Def}\\
\midrule
1.01  & \texttt{((PIQ).=.(-PVQ))}\\
2.33  & \texttt{((PVQVR).=.((PVQ)VR))}\\
3.01  & \texttt{((P*Q).=.-(-(PV-Q))}\\
4.01  & \texttt{((P=Q).=.((PIQ)*(QIP)))}\\
1.2   & \texttt{((AVA)IA)}\\
1.3   & \texttt{(BI(AVB))}\\
1.4   & \texttt{((AVB)I(BVA))}\\
1.5   & \texttt{((AV(BVC))I(BV(AVC)))}\\
1.6   & \texttt{((BIC)I((AVB)I(AVC)))}\\
\bottomrule
\end{tabular}
\end{table}

Table~\ref{tab:results} shows the results of my reanimated LT on the
23 theorems attempted. Of these, 16 are proved and 7 are not---results
that are historically consistent with the original LT's known
behavior, given an effort limit of 20,000 IPL machine cycles.

\begin{table}[h]
\centering
\caption{Theorems attempted}
\label{tab:results}
\begin{tabular}{lll}
\toprule
\textbf{Name} & \textbf{Formula} & \textbf{Result} \\
\midrule
2.01 & \texttt{(PI-P)I-P} & Proved \\
2.02 & \texttt{QI(PIQ)} & Proved \\
2.04 & \texttt{(PI(QIR))I(QI(PIR))} & Proved \\
2.05 & \texttt{(QIR)I((PIQ)I(PIR))} & Proved \\
2.06 & \texttt{(PIQ)I((QIR)I(PIR))} & Proved \\
2.07 & \texttt{PI(PVP)} & Proved \\
2.08 & \texttt{PIP} & Proved \\
2.10 & \texttt{-PVP} & Proved \\
2.11 & \texttt{PV-P} & Proved \\
2.12 & \texttt{PI--P} & Proved \\
2.13 & \texttt{PV---P} & Proved \\
2.14 & \texttt{--PIP} & Not proved \\
2.15 & \texttt{(-PIQ)I(-QIP)} & Not proved \\
2.20 & \texttt{PI(PVQ)} & Proved \\
2.21 & \texttt{-PI(PIQ)} & Not proved \\
2.24 & \texttt{PI(-PVQ)} & Proved \\
3.13 & \texttt{(-(P*Q))I(-PV-Q)} & Not proved \\
3.14 & \texttt{(-PV-Q)I(-(P*Q))} & Not proved \\
3.24 & \texttt{-(P*-P)} & Not proved \\
4.13 & \texttt{P=--P} & Not proved \\
4.20 & \texttt{P=P} & Proved \\
4.24 & \texttt{P=(P*P)} & Proved \\
4.25 & \texttt{P=(PVP)} & Proved \\
\bottomrule
\end{tabular}
\end{table}

A representative proof trace is shown below. Here LT proves Theorem
2.01, \texttt{(PI-P)I-P}:

\begin{lstlisting}
2.01    (PI-P)I-P

PROOF FOUND.

    GIVEN           *1.2  (AVA)IA
    SUBSTITUTION    .0    (PI-P)IP
    SUBLEVEL REPL   2.01  (PI-P)I-P
    Q.E.D.

EFFORT      LIMIT 20000   ACTUAL 5579
SUBPROBLEMS LIMIT 50      ACTUAL 1
SUBSTITUTIONS LIMIT 50    ACTUAL 2
\end{lstlisting}

The proof begins from Axiom~*1.2, \texttt{(AVA)IA}, substitutes
\texttt{-P} for \texttt{A} to obtain \texttt{(-PV-P)I-P}, then applies
sublevel replacement (using the fact that \texttt{P} implies
\texttt{PVP}) to transform the antecedent, yielding the desired
theorem. Appendix 1 depicts a more complex proof (4.25), and the
original output, including both of these proofs, is at: \cite{runoutput}.

\section{Debugging: Solo Effort, Then an AI Collaboration}
\label{sec:debugging}

\subsection{The Solo Phase}

The great majority of the development and debugging work on this
project was done without LLM assistance. Building the interpreter,
transcribing the code, understanding the IPL-V manual, and working
through the first phase of bugs---crashes, unimplemented J-functions,
stack underflows, register save/restore errors---was a solo effort
spanning many months. These bugs were frustrating but tractable: a hard
crash is almost always traceable to some error in the run-up to the
crash, and a human programmer working alone can identify and fix them.

Over this period, I developed a strong sense that the IPL-V
interpreter basically worked---otherwise LT would never have gotten
through even a dozen instructions! And that the LT transcription was
correct. Therefore, almost all remaining errors were likely in the
J-functions, which I had written in Lisp from their descriptions in
the Newell et al.\ manual. My experience was that the manual's
J-function descriptions were sometimes vague in subtle ways. This had
bitten me before, so I was particularly wary of the most complex
J-functions: the generators (J17--J19, J100) and the list manipulation
functions (J74 and many others).

\subsection{The Wall---and How LLMs Broke Through It}

Unlike conventional programs where correctness can be verified
stepwise, LT is a heuristic search program whose behavior emerges from
many interacting procedures.  Once the interpreter was stable enough
that LT ran without crashing, the character of the bugs changed
entirely. The output was now thousands of lines of execution
trace---machine cycles, register states, list operations---that
\textit{might} be correct. LT was proving trivial theorems (those
requiring substitution only), but it would not prove simple theorems
that it should have done trivially, get into runaway recursions,
fail to recognize when a proof had been found, and exhibit other
anomalous behaviors short of crashing. The program was heuristic; it
was supposed to explore and backtrack. Distinguishing a correct
backtrack from one caused by a subtle bug required understanding both
the IPL-V machine semantics and LT's high-level proof strategy
simultaneously, while tracking state across tens of thousands of
interpreted machine cycles.

This experience had an uncanny resonance with the original developers'
accounts. In an interview conducted by Pamela McCorduck for her
pioneering history of AI~\cite{mccorduck2004machines}, Cliff Shaw
describes the particular difficulty of debugging these early AIs,
noting that most daunting phase of troubleshooting occurred during the
latter stages of development, when the system was beyond crashes and
would produce behavior that appeared superficially reasonable, leaving
the developers in a quandary as to whether the results were valid
consequences of the heuristic logic or actual bugs. This created a
profound diagnostic challenge where the impact of code revisions was
difficult to track because the system's memory had become a churning
and impenetrable web of interconnected lists. Unlike structural
failures or crashes, where the error is typically proximal to the
point of failure, these latter anomalies were detached from any
obvious source, making it nearly impossible to distinguish between an
intended emergent property and a hidden flaw.\cite{cmuarchiveshawinterview}

Sixty years later, with the same family of programs, I faced
the identical challenge. At this point progress had effectively
stalled. The cognitive load of holding the machine state in my head
across thousands of trace lines, while simultaneously questioning
whether each line reflected correct or incorrect behavior, exceeded
what I could sustain.

This is where I turned to large language models---specifically,
Google's Gemini (via the AntiGravity interface) and Anthropic's Claude
(via its terminal-based coding interface, Claude Code). They were not
involved in the earlier phases of the project; they entered at the
point where the nature of the debugging task shifted from crash
analysis to the far more demanding problem of debugging non-fatal
mis-behaviors.

\subsection{Teaching a Dead Language to an LLM}

IPL-V is an essentially dead language: the manual is poorly OCR'd in
places and is sometimes obscure even when legible. The language itself
is unlike any modern programming language, and there is no Stack
Overflow, no GitHub corpus, no community of practitioners for the LLMs
to have learned from. Few people remain who have direct experience
programming in IPL-V.  The LLMs' primary source of information about
IPL-V was my own Lisp code---which, of course, still contained bugs.

This created a fascinating epistemological bootstrapping problem. The
LLMs were inferring IPL-V semantics from my interpreter, sometimes
correctly and sometimes not. When the interpreter had bugs, the LLMs
would occasionally learn incorrect semantics and propagate those
errors in their suggestions. Untangling this required constant
vigilance about what was established knowledge versus what was
inferred from possibly-buggy code. Programming with LLMs always
involves a degree of project teaching, but in this case I also had to
teach the language itself---a qualitatively different challenge.

One early difficulty was that an LLM would sometimes try to debug the
LT program rather than my interpreter. We had to maintain the
methodological assumption that the LT code was correct (even though we
did not entirely know this), and focus debugging effort on the
interpreter and especially the J-functions. Without this discipline,
every anomaly would have opened many lines of investigation, making
progress impossible.

\subsection{The J74 Bug: Simon's J's to the Rescue}

What the LLMs excelled at was tenacity. They were willing to read
through thousands of lines of traces and dumps, insert breakpoints at
hundreds of places, and maintain the kind of tedious but precise
state-tracking demanded by this task. This combination of exhaustive
attention and domain comprehension is, ironically, precisely where
current LLMs excel.

Gemini found the first significant issue after several hours of LLM
time spread over several days. As predicted, the problem was in a
J-function. This made substantial progress: now LT was generating
proofs, but was not stopping when a proof was found. After literally
another several hours of LLM time, Claude isolated the second problem
in J74---the list copying function. But we did not know how to fix it,
because both Claude and I agreed that my code appeared to match the
manual's description. There was a clear problem, but the specification
seemed to support my implementation. Based on Claude's explorations, I
developed a hypothesis about what was wrong.

This is where the project took its most interesting turn.  We had
access to ``Simon's J's'' from the Computer History
Museum~\cite{bochannek2012simonsjs}---a set of original IPL-V punched
cards from 1962.  These cards represent the production implementation
of the J-functions at the very moment IPL-V was being standardized for
wider use.  By this point I had taught Claude enough IPL-V that it was
able to read the original 1962 code and compare it with my Lisp
implementation of J74 in an effort to validate my hypothesis of the
problem. The result: they differed in exactly the way I had
hypothesized, and a one-line patch to my J74 implementation made LT
run completely correctly!

This was a remarkable moment: an AI system, having learned a dead
programming language primarily from a modern interpreter that it had
helped debug, was now able to read original 1962 source code and use
it to identify and fix a subtle bug in the heart of the
interpreter. It was a collaboration across sixty years---Simon's
original card deck, preserved by the Computer History Museum and
transcribed by its volunteers, provided the ground truth that neither
the manual's prose description, nor my interpretation of it, nor the
LLM's inference from my code, had been able to supply.

\subsection{Roles of the Collaborators}

The human and LLM collaborators on this project played complementary
roles. The human contributors---particularly Anthony Hay, Arthur
Schwarz, Leigh Klotz, and David M.\ Berry---provided code
transcription, error detection, and sustained intellectual
encouragement through weekly progress updates. Rupert Lane's gridlock
software caught transcription errors that had survived multiple rounds
of human review. These contributions were essential and could not have
been replaced by LLMs.

The LLMs, for their part, filled a role that no human collaborator was
available to fill: the grinding, hour-after-hour work of reading
thousands of lines of execution traces, inserting diagnostic
instrumentation, and tracking machine state across tens of thousands
of cycles to localize subtle bugs. Both Gemini and Claude made
substantive debugging contributions. In the end, I settled on working
primarily with Claude, not because of a significant capability
difference, but because it was easier to maintain a single long
session with accumulated context than to switch between
systems. Interaction histories with both LLMs are haphazardly preserved
in the project's repo on GitHub.

\section{Reflections: IPL, Lisp, and the Arc of AI}
\label{sec:reflections}

I have been programming in Lisp for fifty years. Building an IPL-V
interpreter taught me that Newell, Simon, and Shaw had, by the late
1950s, already invented essentially every conceptual element that
makes Lisp powerful: list manipulation, symbol processing, the idea
that programs are themselves lists, dynamic memory allocation,
recursion, higher-order functions, generators, and even the
possibility of runtime self-modification (which, to be perfectly
honest, all machine-level languages have -- a capability lost in most
higher level languages, except for Lisp). What IPL lacks is the
elegant, lambda-calculus based homoiconic syntax---the elegant
unification of code and data in a single notation that makes Lisp
programs manipulable by Lisp programs.  IPL-V and Lisp coexisted for
approximately five years, but Lisp was so much simpler and more
expressive that it rapidly supplanted its ancestor. (And, of course,
even early Lisps had automatic garbage collection, also invented by
McCarthy (!), which makes programming tolerable, albeit at the cost
of some predictability in terms of performance.)

There is a personal dimension to this realization that bears on the
nature of the project itself. I studied under both Newell and Simon at
Carnegie Mellon University in the 1980s, where my work focused on
modeling the cognition of scientists and engineers. IPL and LT were
long since history by that time; I unfortunately never discussed
either with either Simon or Newell---we barely learned about these
pioneering efforts in our classes. But the intellectual values they
instilled---rigor about mechanism, respect for the complexity of
cognition, and the conviction that thinking could be understood
computationally---shaped the rest of my career. Now, at the end of
that career, I find myself studying Newell, Simon, and Shaw
\textit{as} scientist-engineers: reverse-engineering their design
decisions, reconstructing their reasoning, and trying to understand
how they invented AI and cognitive science. The subject of my
training---the cognition of scientists and engineers---and its object
have folded into one another.

\section{Contribution and Related Work}
\label{sec:related}

This paper makes three contributions:

\begin{enumerate}
\item A working interpreter for IPL-V implemented in Common Lisp.
\item The successful execution of the Logic Theorist from Stefferud’s IPL-V source listing, likely the first such execution in over fifty years.
\item A demonstration of executable reconstruction as a method for studying early AI systems.
\end{enumerate}

The reanimation of historical software systems has gained momentum in
recent years. Most directly related is the ELIZA reanimation by Lane
et al.~\cite{lane2025eliza}, in which a team including the present
author restored Weizenbaum's original ELIZA to operation on a
reconstructed CTSS running on an emulated IBM~7094. That project
reconstructed the complete original hardware and operating system
stack; the present work instead emulates the abstract machine
specification in a modern language.

Reimplementing the Logic Theorist's \textit{algorithms} in a modern
language is a relatively straightforward exercise---essentially an
introductory AI course project, and one that has surely been done many
times. A 1968 paper, for example, describes an implementation of LT in
LISP~1.5~\cite{millstein1968logic}. The present work is a
fundamentally different undertaking: rather than rewriting LT's logic
in a modern language, it constructs a faithful emulator of the IPL-V
abstract machine and runs the original code, transcribed directly from
a 1963 technical report. The distinction is significant: a Lisp
reimplementation demonstrates that LT's algorithms can be expressed in
Lisp (which is unsurprising), while the present work demonstrates that
the original IPL-V code, as written by Newell, Simon, and Shaw's team,
functions as described, and, moreover, enables us to perform
experiments on the real, working system of machine and program.

\section{Next Steps}
\label{sec:next}

The IPL-V interpreter developed for this project is not specific to
the Logic Theorist. It implements the general IPL-V abstract machine
and can, in principle, run any IPL-V program. I am in the process of
building a corpus of early IPL-V AI programs. The most significant
targets are the General Problem Solver (GPS)~\cite{newell1959gps} and
EPAM (Elementary Perceiver and
Memorizer)~\cite{feigenbaum1961epam}. EPAM source code is available in
the Feigenbaum archives at Stanford, and archival work to locate
complete source code for GPS is in progress. Now that a proven IPL-V
interpreter exists, running these programs should require
substantially less effort than the initial construction of the
interpreter itself.

The interpreter, the transcribed LT code, complete outputs as
described in this paper, as well as partial LLM interaction histories
are open sourced at \url{https://github.com/jeffshrager/IPL-V}.

\section*{Acknowledgements}

The following people (listed in no particular order) contributed to
this project in ways ranging from code transcription to weekly
encouragement: Anthony Hay, Arthur Schwarz, Leigh Klotz, David
M.\ Berry, Rupert Lane, Amy Majczyk, Emily Davis, Ethan Ableman, and
Paul McJones. I am particularly grateful to Hay, Schwarz, Klotz, and
Berry, who put up with weekly progress updates throughout the project
and responded with encouragement that was always helpful and sometimes
technically substantive. Rupert Lane's gridlock software caught
several transcription errors that had survived multiple rounds of
human review.

I also thank the Computer History Museum for preserving and making
available ``Simon's J's,'' the original 1962 IPL-V J-function card
deck, which proved essential during debugging.

Finally, I want to acknowledge Allen Newell and Herbert Simon, under
whom I studied at CMU in the 1980s. The subject of that
training---the cognition of scientists and engineers---has, in this
project, become inseparable from its object.

{\footnotesize This paper was written with the aid of various LLMs, but all concepts
and final editorial decisions were the author's alone.}

\clearpage 
\onecolumn 

\appendix

\section{Example Complex Logic Theorist Proof}

To illustrate a complex proof generated by the reconstructed Logic Theorist, the following derivation corresponds to theorem *4.25 of \textit{Principia Mathematica}:

\[ P = (P \lor P) \]

This states the idempotence of disjunction. In the LT representation, the equivalence operator is defined using implication and conjunction:

\[ (P = Q) \equiv ((P \rightarrow Q) \land (Q \rightarrow P)) \]

The Logic Theorist proof trace proceeds as follows:

\begin{enumerate}

\item \textbf{Expand the definition of equivalence:}
LT begins by reducing the goal to a conjunction of two implications.
\begin{quote}
\begin{verbatim}
4.25    P=(PVP)
.0   (PI(PVP))*((PVP)IP)  4.01  ,0 REPLACEMENT
\end{verbatim}
\end{quote}
\textit{Logic form:}
\begin{itemize}
    \item[] $P = (P \lor P) \Rightarrow (P \rightarrow (P \lor P)) \land ((P \lor P) \rightarrow P)$ 
\end{itemize}

\item \textbf{Identify the first implication:}
The system identifies a forward-chaining path to establish the first half of the conjunction.
\begin{quote}
\begin{verbatim}
PI(PVP)    4.20  ,0 FORWARD CHAINING
\end{verbatim}
\end{quote}
\textit{Logic form:} 
\begin{itemize}
    \item[] $P \rightarrow (P \lor P)$
\end{itemize}

\item \textbf{Find the proof via substitution:}
LT utilizes Theorem *2.20 from its memory as the basis for substitution. \textit{This was not a given axiom, but a previously proved theorem that it remembered and applied here; See comment below.}
\begin{quote}
\begin{verbatim}
GIVEN      2.20      AI(AVB)
SUBSTITUTION  .0    PI(PVP)
\end{verbatim}
\end{quote}
\textit{Logic form:} 
\begin{itemize}
    \item[] The general theorem $A \rightarrow (A \lor B)$ is retrieved
    \item[] Substituting $A := P$ and $B := P$ yields $P \rightarrow (P \lor P)$
\end{itemize}

\item \textbf{Complete the equivalence:}
LT completes the chain by linking the proved implications back to the original goal.
\begin{quote}
\begin{verbatim}
GIVEN       4.20      A=A
FORWARD CHAINING    4.25      P=(PVP)
Q.E.D.0
\end{verbatim}
\end{quote}
\textit{Logic form:} 
\begin{itemize}
    \item[] Because both $P \rightarrow (P \lor P)$ and $(P \lor P) \rightarrow P$ are established, the equivalence $P = (P \lor P)$ holds.
\end{itemize}

\end{enumerate}

\textbf{Performance Metrics:}
\begin{quote}
\begin{verbatim}
EFFORT              LIMIT 20000         ACTUAL 8727
SUBPROBLEMS         LIMIT 50            ACTUAL 2
SUBSTITUTIONS       LIMIT 50            ACTUAL 3
\end{verbatim}
\end{quote}

\medskip

\noindent The successful execution of more complex derivations, such as Theorem
4.25, provides a striking demonstration of the Logic Theorist’s
capacity for a primitive form of machine learning. By treating
previously proved theorems, such as 2.20, as established 'givens' for
subsequent search paths, the system effectively bootstraps its own
knowledge base, mirroring the cumulative nature of algebraic reasoning.

\end{document}